\documentclass[conference]{IEEEtran}
\IEEEoverridecommandlockouts

\usepackage{cite}
\usepackage{amsmath,amssymb,amsfonts}
\usepackage{graphicx}
\usepackage{booktabs}
\usepackage{multirow}
\usepackage[hidelinks]{hyperref}
\hypersetup{
  pdftitle={Localize, Don't Beautify: Client-Side Control of Image-Editing APIs for Cosmetic Surgery Previews},
  pdfauthor={Sukhrobbek Ilyosbekov}
}

\graphicspath{{figures/generated/}{figures/static/}}

\IfFileExists{tables/numbers.tex}{
\newcommand{\NumScoredEdits}{151}
\newcommand{\NumAttemptedEdits}{162}
\newcommand{\NumExcludedEdits}{11}

\newcommand{\NumFacesScored}{16}
\newcommand{\NumPrimaryFaces}{15}
\newcommand{\NumLocalizationFaces}{15}
\newcommand{\MedPairedLocGain}{0.315}
\newcommand{\PairedLocGainLow}{0.279}
\newcommand{\PairedLocGainHigh}{0.366}
\newcommand{\MedPairedOffReduction}{3.10}
\newcommand{\PairedOffReductionLow}{2.91}
\newcommand{\PairedOffReductionHigh}{3.37}
\newcommand{\IdentityMinAll}{0.460}

\newcommand{\NumBelowFloor}{8}
\newcommand{\MedLocPromptOnly}{0.682}
\newcommand{\LocPromptOnlyMin}{0.489}
\newcommand{\LocPromptOnlyMax}{1.000}
\newcommand{\MedIdentityPromptOnly}{0.803}
\newcommand{\MedTgtPromptOnly}{6.6}
\newcommand{\MedOffPromptOnly}{3.14}
\newcommand{\MedLocComposite}{0.989}
\newcommand{\LocCompositeMin}{0.970}
\newcommand{\LocCompositeMax}{0.999}
\newcommand{\MedIdentityComposite}{0.820}
\newcommand{\MedTgtComposite}{6.1}
\newcommand{\MedOffComposite}{0.04}
\newcommand{\NumBelowFloorSunburst}{6}
\newcommand{\MedIdentityPromptOnlySunburst}{0.761}
\newcommand{\MedTgtFaceliftPromptOnlySunburst}{6.8}
\newcommand{\MedTgtRhinoPromptOnlySunburst}{8.5}
\newcommand{\NumBelowFloorFlare}{2}
\newcommand{\MedIdentityPromptOnlyFlare}{0.777}

\newcommand{\MedIdentityPromptOnlyNanoBanana}{0.842}
\newcommand{\MedTgtFaceliftPromptOnlyNanoBanana}{3.8}

\newcommand{\MedIdentityPromptOnlySeedream}{0.803}
\newcommand{\MedTgtFaceliftPromptOnlySeedream}{5.0}

\newcommand{\MedIdentityPromptOnlyFlux}{0.862}
\newcommand{\MedTgtFaceliftPromptOnlyFlux}{0.9}
\newcommand{\MedTgtRhinoPromptOnlyFlux}{7.3}
\newcommand{\NumChained}{2}
\newcommand{\ChainedIdentityLow}{0.665}
\newcommand{\ChainedIdentityHigh}{0.691}
\newcommand{\ChainedLocLow}{0.964}
\newcommand{\ChainedLocHigh}{0.968}
\newcommand{\MedLatency}{48}
\newcommand{\MedCost}{0.053}
}{}

\begin{document}

\title{Localize, Don't Beautify: Client-Side Control of Image-Editing APIs for Cosmetic Surgery Previews}

\author{\IEEEauthorblockN{Sukhrobbek Ilyosbekov}
\IEEEauthorblockA{Northeastern University\\
ilyosbekov.s@northeastern.edu}}

\maketitle

\begin{abstract}
Ask a commercial image editor to preview a cosmetic procedure and it will often
change more of the face than the request names: a jawline edit can also smooth
the forehead or alter the eyes. Existing methods for confining an edit to one
region require access to the model's internals, which a public editing API does
not expose. We ask how much control is possible from the client side alone. In a
pilot benchmark, five current commercial editing configurations perform
facelift-style jaw--neck and rhinoplasty edits on synthetic faces under two
levels of client-side control: the prompt alone, and a masked composite that
cuts the edited region out of the response and pastes it back onto the original
photograph through a landmark-derived mask. Of \NumAttemptedEdits{} attempted
edits, \NumScoredEdits{} could be scored. ArcFace cosine measures identity
preservation; a CIELAB pixel-change ratio measures how much change lands inside
the requested region rather than a protected facial zone. On
\NumLocalizationFaces{} faces, the masked composite improved localization over
the paired prompt-only output by a median of \MedPairedLocGain{} (95\%
face-clustered bootstrap interval \PairedLocGainLow--\PairedLocGainHigh) while
retaining most of the target-region change. Editors differed in edit strength
and identity retention, and one returned near-copies for the facelift request.
This is a study of control, not clinical accuracy: the faces are synthetic,
no surgeons rated the outputs, and each condition was generated once. Within
that scope, keeping a surgical preview inside its intended region needs no access
to the model; a mask and composite on the client enforce it across every editor
tested, at low provider cost.

\end{abstract}


\section{Introduction}
\label{sec:intro}

\begin{figure*}[t]
  \centering
  \includegraphics[width=0.88\textwidth]{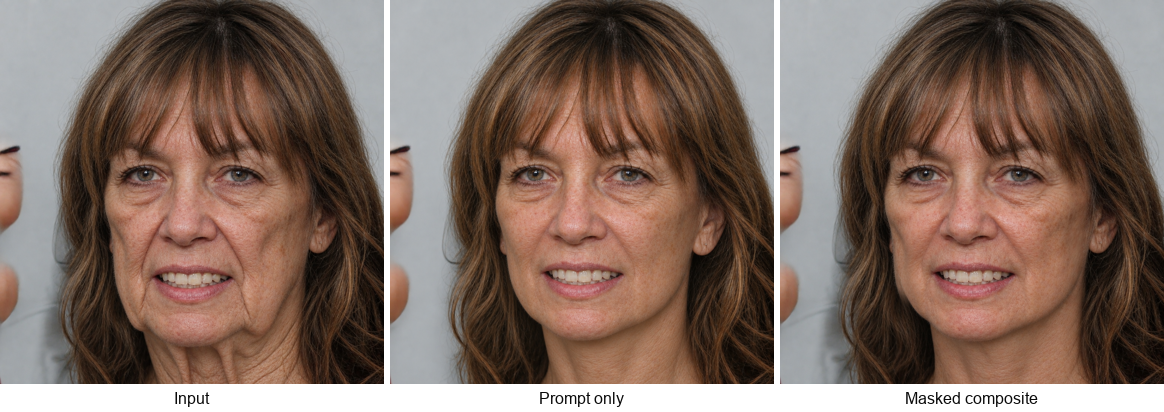}
  \caption{A facelift-style jaw--neck edit of one synthetic face (GPT Image~2.6
  Sunburst). The prompt-only output (center) also smooths the forehead and the
  skin around the eyes; the masked composite (right) retains the model output
  only inside the blend mask.}
  \label{fig:teaser}
\end{figure*}

Surgeons have used simulated outcomes in patient consultations for decades.
Computer imaging has been associated with improved rhinoplasty patient
satisfaction~\cite{sharp2002computerimaging}, and how closely a 3D simulation
matches the eventual outcome predicts how satisfied the patient will
be~\cite{yamamichi2025crisalix}. The same literature warns that an over-idealized
preview sets expectations no operation can meet~\cite{agarwal2007morph}. Producing
these previews today requires either manual morphing skill or dedicated 3D
systems; meanwhile, general-purpose image-editing APIs, i.e., commercial editing
models reachable only through a public endpoint, can perform instruction-guided
photo edits. The natural question is whether such a service can produce a
controlled cosmetic-surgery preview when neither its weights nor its inference
process is accessible.

In preliminary tests, a request for a specific visible change, such as jawline
and neck repositioning or dorsal-hump reduction, also altered skin texture,
wrinkles, and other facial regions. Off-target change of this kind can make the output less
recognizable as the same person, and it shows the patient cosmetic effects
unrelated to the requested procedure. The
training-free localization literature knows how to confine
edits~\cite{avrahami2022blended,couairon2023diffedit}. Every such method, however,
needs the model's latents, attention, or sampling loop, and a public editing
API exposes none of them.

We therefore evaluate how much off-target change can be prevented on the
client, i.e., the caller's side of the API, before a request is sent or after
the response returns.
The pilot benchmark covers five current editing configurations (GPT Image~2.6
in its Sunburst and Flare tiers, Nano Banana~2.1, Seedream~5.0 Pro, and FLUX~3),
two requested edit types, and synthetic faces, so no real person is depicted;
each matrix cell contains one generated output.
Our contributions are:

\begin{itemize}
  \item \textbf{A model-agnostic client-side control.} Over the same black-box
  APIs we compare prompt-only editing with a client-side masked composite that
  cuts the model's edit out along a landmark-derived region mask and pastes it
  back onto the original photograph (with alignment, feathered blending, and
  pose gating). The composite adapts the final step of blended-diffusion editing
  to a setting where nothing inside the model is reachable
  (Sec.~\ref{sec:method}).
  \item \textbf{A two-part diagnostic protocol.} ArcFace identity is paired with
  a CIELAB region-localization ratio and its target/off-target components. The
  metrics detect identity loss, off-target pixel change, and near-copies, but do
  not measure clinical plausibility (Sec.~\ref{sec:protocol}).
  \item \textbf{A pilot benchmark.} Client-side compositing enforces off-mask
  preservation across the tested editors while retaining most on-target pixel
  change. The editors differ in edit strength and identity retention, and the
  metrics expose one editor that returns near-copies for the facelift request. A
  two-order, single-face chain is reported only as a feasibility probe
  (Sec.~\ref{sec:results}).
\end{itemize}

To our knowledge, this is the first controlled comparison of client-side region
confinement across multiple commercial, black-box editing APIs for
cosmetic-surgery previews. What differs here is the setting. Bespoke surgical
simulators and composited rhinoplasty systems
exist~\cite{knoedler2024rhinoplastygan,agarwal2026envisage}, but they assume a
model the operator hosts, and image-editing APIs have been benchmarked in other
domains~\cite{qian2025giebench,gptimgeval2025} without control as the
manipulated variable. When the editor is a black-box API, no
internals-based localization method applies, so whatever control exists must
live on the client. Measuring how much that recovers is our contribution; we
propose no new compositing algorithm, clinical outcome model, or clinical
validation study (Sec.~\ref{sec:related}).\footnote{Analysis code and paper source:
\url{https://github.com/suxrobGM/localize-dont-beautify}}

\section{Related Work}
\label{sec:related}

\textbf{Surgical outcome simulation.} Preoperative visualization predates
generative models by decades: manual computer morphing measurably improves
rhinoplasty patient satisfaction~\cite{sharp2002computerimaging}, the risks of
setting expectations too high are equally well
documented~\cite{agarwal2007morph},
and 3D simulation accuracy correlates with postoperative
satisfaction~\cite{yamamichi2025crisalix,threedvr2026aesthetic}. Learning-based
successors train bespoke models per procedure: a GAN trained on 3{,}030 paired
rhinoplasty photographs whose outputs raters could not distinguish from real
post-ops~\cite{knoedler2024rhinoplastygan}, appearance predictors for
blepharoptosis~\cite{blepharoptosis2022poap,huang2024ptosisdiffusion}, and
biomechanics-driven craniofacial simulation~\cite{fang2022acmtnet}; a recent
systematic review surveys the area~\cite{stephanian2024fpsamreview}. Closest to us,
Envisage~\cite{agarwal2026envisage} combines a local FLUX-based inpainter with
MediaPipe masks, hard compositing, and a mask-decomposed score built on masked
LPIPS~\cite{zhang2018lpips}. It evaluates rhinoplasty on 211 cases and shows why
full-face ArcFace scores are confounded after compositing. We address a
different systems question: Envisage operates an open-weight inpainter it
hosts itself, whereas our editors are black-box APIs, and neither its
pipeline nor any internals-based localization method applies to them. We compare
five such configurations on two requested edit types with control itself as the
manipulated variable. Our study is smaller: we do not reproduce Envisage's
clinical presets, mask-decomposed fidelity score, sample size, or
ground-truth-relative analysis, which limits direct quality comparisons. Lim et
al.~\cite{lim2023generativecosmetic} prompt text-to-image tools about the same
procedures but generate generic faces rather than editing a patient's photo.

\textbf{Training-free localized editing.} Methods that confine an edit without
fine-tuning uniformly reach inside a diffusion model~\cite{rombach2022ldm}:
SDEdit, RePaint, and DiffEdit steer the sampling
loop~\cite{meng2022sdedit,lugmayr2022repaint,couairon2023diffedit},
Blended (Latent) Diffusion blends masked latents at every denoising
step~\cite{avrahami2022blended,avrahami2023blendedlatent}, and Prompt-to-Prompt
and Plug-and-Play inject attention or
features~\cite{hertz2023prompt2prompt,tumanyan2023plugandplay}. A black-box API
exposes none of those surfaces. Our masked composite is deliberately the
post-hoc remnant of this family, the final pixel-space blend applied after the
API returns, and we measure how much control that remnant still provides.

\textbf{Benchmarking instruction-guided editors.} General-domain benchmarks are
plentiful~\cite{brooks2023instructpix2pix,zhang2023magicbrush,wang2023editbench,
sheynin2024emuedit,ku2024imagenhub,ma2024i2ebench,hui2025hqedit,liu2025step1xedit},
and commercial editing APIs are now benchmarked
routinely~\cite{gptimgeval2025,qian2025giebench}. Off-target measurement also has
precedent: GIE-Bench scores object-masked preservation and finds that GPT-Image
over-edits irrelevant regions~\cite{qian2025giebench}, and ALE-Bench quantifies
attribute leakage outside the edit target~\cite{alebench2024}. These benchmarks
do not address cosmetic-surgery preview editing. Our pilot pairs an identity
diagnostic with facial target/keep-zone pixel change, but unlike clinically
validated outcome studies it has no surgeon ratings or input-relative
postoperative accuracy measure. ArcFace cosine itself is standard in face
editing~\cite{deng2019arcface}; neither it nor the CIELAB ratio is claimed as a
new or clinically validated metric.

\textbf{Face analysis.} The pipeline builds on MediaPipe's face
mesh~\cite{lugaresi2019mediapipe,grishchenko2020attentionmesh} with InsightFace's
SCRFD detection and landmarks as the profile fallback~\cite{guo2022scrfd}, and on
CIELAB color difference~\cite{sharma2005ciede2000} for the regional metric.

\section{Method}
\label{sec:method}

We treat each editing model as a black box behind a remote API: it accepts an
image and a text instruction and returns an edited image. No weights, latents,
attention maps, or sampling loop are exposed, so the training-free localization
methods surveyed in Sec.~\ref{sec:related} do not apply. Everything we add runs on
the client, before the request or after the response.

\subsection{The two controls}
\label{sec:ladder}

A procedure specification consists of a target region $r$ (jaw and neck for the
facelift-style request, nose for rhinoplasty), a positive instruction describing
the visible anatomical change, and a list of changes to avoid. We compare two
levels of client-side control, holding faces, model settings, and prompts fixed.
A common seed is supplied to every model, but only the Nano Banana~2.1 schema has
a seed field; the GPT Image~2.6, Seedream~5.0 Pro, and FLUX~3 schemas ignore it:

\begin{itemize}
  \item \textbf{Prompt only.} The photo and instruction go to the endpoint
  unmodified and the response is used as-is. This measures what the model does when
  instructed to edit one region.
  \item \textbf{Masked composite.} The model still edits the whole image, but we
  discard everything outside $r$: the response is aligned back onto the input and
  composited through a feathered region mask (Sec.~\ref{sec:composite}), so pixels
  outside the mask remain identical to the input by construction.
\end{itemize}

The control is the experimental variable, not a new editing algorithm: pixel-space
blending is the final step of Blended Diffusion~\cite{avrahami2022blended} and of
Envisage's hard composite~\cite{agarwal2026envisage}. What we study is how much of
the localized-editing problem this post-hoc adaptation solves when nothing inside
the model can be touched.

\subsection{Landmarks and region masks}
\label{sec:masks}

Region masks derive from MediaPipe's 478-point face
mesh~\cite{lugaresi2019mediapipe,grishchenko2020attentionmesh}. Each region is the
convex hull of a landmark subset: the face splits at the lower-eyelid line into an
upper-face zone (forehead, brows, and eyes, which a facelift must not change) and a
jaw--neck zone (cheeks, nasolabial folds, jowls, jawline, chin). The mesh has no
neck landmarks, so the jaw--neck hull extends below the chin by a fixed fraction of
face height. The nose region is a perimeter hull dilated by a margin proportional
to the inter-ocular distance, which also scales every feather and erosion radius so
the masks are resolution-independent.

MediaPipe finds no face on true profile views. For the nose region we fall back to
InsightFace's SCRFD detector with its 106-point landmark
set~\cite{guo2022scrfd}, scaled by the detection box; the remaining regions are
defined only by the frontal mesh and are pose-gated instead (below).

\subsection{Masked compositing and pose gating}
\label{sec:composite}

Editing APIs routinely shift, rescale, or re-crop the head, so pasting the raw
response through a mask produces double contours. We first align the edited image
onto the input with a similarity warp estimated from stable landmark anchors, then
blend: $\mathrm{out} = x\,(1-w) + e\,w$, where $x$ is the input, $e$ the aligned
edit, and $w$ a per-pixel weight from the feathered mask.

The composite mask is shaped so its boundary never crosses visually busy texture.
Relative to the metric region, we erode it off the face silhouette, clip it medial
of the ears so the seam avoids the sideburns, and extend it further down the neck
so a narrowed jawline blends into shadow or clothing. The feather is a wide
smoothstep ramp over the signed distance to the mask boundary rather than a
Gaussian blur. All widths are inter-ocular fractions.

Compositing is skipped when the estimated yaw exceeds $0.35$. In early profile
rhinoplasty runs the frontal-only hulls landed on the wrong part of the face and
pasted a translucent ``ghost nose'' onto the cheek; gating on pose returns the
unmodified edit for those inputs rather than a corrupted one.

\subsection{Prompt guardrails}
\label{sec:guardrails}

The dominant failure mode we observed is not refusal but \emph{generic
beautification}: smoothed skin, brightened teeth, and subtly enlarged eyes. Every
request therefore carries three fixed instruction fragments in addition to the
procedure text. An avoidance fragment lists the beautification behaviors
(``beauty filter, skin smoothing,
airbrushing, wrinkle removal, \ldots{} different person, identity change, face
swap''), a positive fragment anchors realism and framing (``photorealistic, same
person, identity preserved, natural skin texture, identical camera framing''), and
an intensity fragment scopes the change to anatomy (``change only facial tissue and
contour; keep skin tone, color, exposure, and lighting identical to the input'').
Where a separate negative-prompt field exists, the avoidance fragment is supplied
there; otherwise it is appended to the main instruction. The prompts request a
visible anatomical correction because target and off-target changes are reported
together (Sec.~\ref{sec:protocol}).

\section{Evaluation Protocol}
\label{sec:protocol}

An edit can change too much outside the requested region or too little inside it.
Our automated protocol records both directions, but it is a systems diagnostic:
it does not establish whether an edit is anatomically or clinically plausible.

\subsection{Identity}

We compute the cosine similarity between ArcFace~\cite{deng2019arcface}
embeddings of the largest detected face in the input and in the edit. We plot
$0.6$ as a heuristic reference threshold, not a clinically calibrated acceptance
criterion; verification thresholds depend on implementation, preprocessing, and
the desired false-match rate. A high score can also mean the model did not perform
the edit, so identity is read with the regional-change scores below.

As Agarwal and Bhrany~\cite{agarwal2026envisage} note, a full-face identity score
on a composited output is partly determined by pixels copied from the input. We
therefore treat prompt-only identity as a model-level diagnostic and composite
identity as a system-level property. Neither score measures surgical accuracy.

\subsection{Regional change and localization}

After aligning the edit onto the input, we convert both to CIELAB and take the
mean $\Delta E_{76}$ color difference inside the requested target region
($\Delta E_{\mathrm{tgt}}$) and inside a disjoint facial keep zone
($\Delta E_{\mathrm{off}}$, the upper-face region minus the target). We use the
simple 1976 distance rather than the perceptually refined
CIEDE2000~\cite{sharma2005ciede2000}; it is a reproducible pixel-change
diagnostic, not a perceptual or anatomical accuracy score. The summary statistic
is
\[
  \mathrm{loc} = \frac{\Delta E_{\mathrm{tgt}}}
                      {\Delta E_{\mathrm{tgt}} + \Delta E_{\mathrm{off}}}.
\]
The ratio is read with its components: low $\Delta E_{\mathrm{tgt}}$ identifies
a near-copy, while high $\Delta E_{\mathrm{off}}$ indicates leakage into the
measured keep zone. Related preservation metrics appear in general-domain editing
benchmarks~\cite{qian2025giebench,alebench2024,sheynin2024emuedit}.

The ratio has three limitations. First, the keep zone does not cover the entire
image; lower-face, hair, clothing, or background changes may be missed,
particularly for rhinoplasty. Second, the frontal-mesh metric abstains on profile
inputs. Third, compositing copies off-mask pixels by design, so a high ratio on
the composite verifies the implementation's preservation property rather than the
editor's quality.

\subsection{Descriptive uncertainty}

The face, not the generated edit, is the sampling unit. Cell summaries are
medians and ranges because each model--control cell contains fewer than ten faces.
For the primary prompt-only versus masked-composite contrast, we compute paired
differences within face, procedure, and model, then resample faces (10{,}000
replicates, fixed analysis seed) to obtain a percentile interval for the median.
This interval describes the observed face set; it does not repair the small sample
or support population-level clinical inference.

\section{Experimental Setup}
\label{sec:setup}

\begin{table}[t]
  \centering
  \caption{Models in the study. All run behind fal.ai; cost is the provider's
  per-image estimate on November 17, 2026. Aliases may change after that date.}
  \label{tab:models}
  \resizebox{\columnwidth}{!}{
\begin{tabular}{llr}
\toprule
Model & Endpoint & Cost/image (\$) \\
\midrule
FLUX 3 & {\scriptsize\texttt{blackforestlabs/flux-3/edit-image}} & 0.050 \\
GPT Image 2.6 Flare & {\scriptsize\texttt{openai/gpt-image-2.6/flare/edit}} & 0.053 \\
GPT Image 2.6 Sunburst & {\scriptsize\texttt{openai/gpt-image-2.6/sunburst/edit}} & 0.053 \\
Nano Banana 2.1 & {\scriptsize\texttt{google/nano-banana-2.1/edit}} & 0.042 \\
Seedream 5.0 Pro & {\scriptsize\texttt{bytedance/seedream/v5/pro/edit}} & 0.068 \\
\bottomrule
\end{tabular}
}
\end{table}

\textbf{Models.} Table~\ref{tab:models} lists five commercial editing
configurations: OpenAI GPT Image~2.6 in its quality (Sunburst) and fast (Flare)
tiers~\cite{fal2026gptimage26}, Google's Nano Banana~2.1~\cite{fal2026nanobanana21},
ByteDance Seedream~5.0 Pro~\cite{fal2026seedream5pro,seedream4_2025}, and Black
Forest Labs FLUX~3~\cite{fal2026flux3,bfl2025fluxkontext}. These were the newest
edit endpoint of each family available on the gateway at the time of the runs.
All requests were routed through a single API gateway, fal.ai, whose endpoint
aliases and per-image prices Table~\ref{tab:models} records; the underlying
models are served by their respective vendors. Several of these endpoints bill
by token, so the listed prices are per-image estimates. These services are
moving targets, so every run records its date and endpoint alias. The gateway
did not expose immutable revisions for every model.

\textbf{Data.} All faces are synthetic. Eight identities were sampled from a
StyleGAN2 face generator (thispersondoesnotexist.com), and each was turned into
one facelift and one rhinoplasty input by an instruction edit
(GPT Image~2.6 Sunburst) that adds age-related jaw--neck laxity or a
procedure-typical nasal shape at a fixed mild, moderate, or pronounced severity.
The two procedure groups therefore share the same eight underlying identities.
There are no postoperative photographs. One rhinoplasty input defeats the
identity detector on every edit (Sec.~\ref{sec:failures}), so the main analysis
contains \NumPrimaryFaces{} procedure-specific faces. With the chained-experiment
input included, \NumFacesScored{} face inputs contribute scoreable outputs.

Images are normalized to $1024^2$ with padding. No photograph of a real person
was used (Sec.~\ref{sec:ethics}).

\textbf{Protocol.} Runs expand a Cartesian matrix of faces $\times$ models
$\times$ requested edit types $\times$ controls. One output is generated per
cell, so the study does not estimate within-model stochastic variation. The gallery renderer shows attempted cells,
and provider or scoring failures remain in the run records. Each immutable run
directory stores its resolved configuration and git commit; output filenames
encode the full cell key used for scoring. Generation and scoring are decoupled,
so metrics can be recomputed without new API requests.

The reported results come from a designated set of matrix runs executed after the
pipeline was frozen; earlier development runs are retained but not pooled. One
additional experiment applies rhinoplasty and facelift-style edits to one facelift
input in both orders as a feasibility check. The full design attempted
\NumAttemptedEdits{} outputs: \NumScoredEdits{} were scoreable and
\NumExcludedEdits{} were unavailable because the identity detector found no face
in one rhinoplasty input (all of its cells) or in one output. Across
scoreable outputs, median latency is \MedLatency{}\,s and median provider cost is
\$\MedCost{} per image.

The paper's numeric values are generated by an aggregation script from the run
records. Duplicate stems are resolved by keeping the row with the most complete
metric set. The public repository includes the canonical score table and
analysis scripts, but not the input images or full generation pipeline
(Sec.~\ref{sec:limitations}).

\section{Results}
\label{sec:results}

\begin{table*}[t]
  \centering
  \caption{Results over \NumScoredEdits{} scoreable outputs from
  \NumAttemptedEdits{} attempts (\NumPrimaryFaces{} main-analysis faces plus one
  chain input). Medians per cell; identity shows min--max. Loc.\ is the
  diagnostic localization ratio and $\Delta E_{\mathrm{tgt}}$ the mean
  target-region pixel change. Chained rows are single-face feasibility probes.}
  \label{tab:main}
\begin{tabular}{lllrlllrr}
\toprule
Procedure & Model & Control & $n$ & Identity$\uparrow$ & Loc.$\uparrow$ & $\Delta E_{\mathrm{tgt}}$ & Lat.\,(s) & Cost\,(\$) \\
\midrule
Facelift-style jaw--neck & FLUX 3 & Masked composite & 8 & 0.966 {\scriptsize(0.899--0.987)} & 0.978 & 2.3 & 50 & 0.050 \\
Facelift-style jaw--neck & FLUX 3 & Prompt only & 8 & 0.960 {\scriptsize(0.862--0.993)} & 1.000 & 0.9 & 51 & 0.050 \\
Facelift-style jaw--neck & GPT Image 2.6 Flare & Masked composite & 8 & 0.832 {\scriptsize(0.787--0.861)} & 0.988 & 5.4 & 16 & 0.053 \\
Facelift-style jaw--neck & GPT Image 2.6 Flare & Prompt only & 8 & 0.802 {\scriptsize(0.777--0.858)} & 0.606 & 6.3 & 17 & 0.053 \\
Facelift-style jaw--neck & GPT Image 2.6 Sunburst & Masked composite & 8 & 0.829 {\scriptsize(0.790--0.873)} & 0.985 & 5.9 & 56 & 0.053 \\
Facelift-style jaw--neck & GPT Image 2.6 Sunburst & Prompt only & 8 & 0.811 {\scriptsize(0.761--0.842)} & 0.617 & 6.8 & 44 & 0.053 \\
Facelift-style jaw--neck & Nano Banana 2.1 & Masked composite & 8 & 0.904 {\scriptsize(0.820--0.938)} & 0.979 & 3.8 & 28 & 0.042 \\
Facelift-style jaw--neck & Nano Banana 2.1 & Prompt only & 8 & 0.872 {\scriptsize(0.770--0.917)} & 0.537 & 3.8 & 26 & 0.042 \\
Facelift-style jaw--neck & Seedream 5.0 Pro & Masked composite & 8 & 0.842 {\scriptsize(0.745--0.869)} & 0.983 & 4.9 & 99 & 0.068 \\
Facelift-style jaw--neck & Seedream 5.0 Pro & Prompt only & 8 & 0.835 {\scriptsize(0.770--0.884)} & 0.604 & 5.0 & 90 & 0.068 \\
Chained: facelift $\to$ rhino. & GPT Image 2.6 Sunburst & Masked composite & 1 & 0.691 {\scriptsize(0.691--0.691)} & 0.964 & 5.5 & 134 & 0.106 \\
Rhinoplasty & FLUX 3 & Masked composite & 7 & 0.790 {\scriptsize(0.703--0.845)} & 0.998 & 6.7 & 52 & 0.050 \\
Rhinoplasty & FLUX 3 & Prompt only & 7 & 0.813 {\scriptsize(0.729--0.840)} & 0.991 & 7.3 & 54 & 0.050 \\
Rhinoplasty & GPT Image 2.6 Flare & Masked composite & 7 & 0.701 {\scriptsize(0.617--0.751)} & 0.998 & 7.9 & 15 & 0.053 \\
Rhinoplasty & GPT Image 2.6 Flare & Prompt only & 7 & 0.690 {\scriptsize(0.534--0.698)} & 0.734 & 8.2 & 17 & 0.053 \\
Rhinoplasty & GPT Image 2.6 Sunburst & Masked composite & 7 & 0.648 {\scriptsize(0.460--0.723)} & 0.998 & 7.8 & 111 & 0.053 \\
Rhinoplasty & GPT Image 2.6 Sunburst & Prompt only & 7 & 0.591 {\scriptsize(0.479--0.684)} & 0.734 & 8.5 & 125 & 0.053 \\
Rhinoplasty & Nano Banana 2.1 & Masked composite & 7 & 0.829 {\scriptsize(0.788--0.905)} & 0.998 & 6.7 & 25 & 0.042 \\
Rhinoplasty & Nano Banana 2.1 & Prompt only & 7 & 0.793 {\scriptsize(0.706--0.853)} & 0.673 & 6.9 & 26 & 0.042 \\
Rhinoplasty & Seedream 5.0 Pro & Masked composite & 7 & 0.751 {\scriptsize(0.678--0.765)} & 0.998 & 8.5 & 97 & 0.068 \\
Rhinoplasty & Seedream 5.0 Pro & Prompt only & 6 & 0.701 {\scriptsize(0.636--0.802)} & 0.733 & 8.4 & 92 & 0.068 \\
Chained: rhino.\ $\to$ facelift & GPT Image 2.6 Sunburst & Masked composite & 1 & 0.665 {\scriptsize(0.665--0.665)} & 0.968 & 6.9 & 101 & 0.106 \\
\bottomrule
\end{tabular}

\end{table*}

\begin{figure}[t]
  \centering
  \includegraphics[width=\columnwidth]{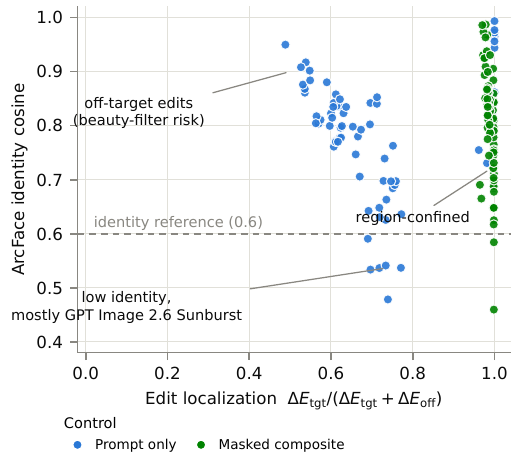}
  \caption{Every scoreable output placed by the two automated diagnostics.
  Compositing produces localization near 1 because it preserves off-mask pixels
  by construction; most prompt-only outputs change more of the measured keep
  zone. Prompt-only points near 1 are near-copies (Sec.~\ref{sec:identity}).
  \NumBelowFloor{} outputs fall below the 0.6 identity reference,
  \NumBelowFloorSunburst{} from GPT Image~2.6 Sunburst.}
  \label{fig:scatter}
\end{figure}

Table~\ref{tab:main} and Fig.~\ref{fig:scatter} summarize the scoreable outputs.
These are repeated model/control observations on \NumPrimaryFaces{}
main-analysis faces, not \NumScoredEdits{} independent patients. We report
descriptive summaries and a face-clustered interval for the primary paired
contrast; no population-level or clinical significance is claimed.

\subsection{Off-target preservation}

Masked compositing produces a median localization ratio of \MedLocComposite{}
(range \LocCompositeMin--\LocCompositeMax) across the five editing
configurations (Fig.~\ref{fig:strips}a). Their prompt-only outputs have a median
of \MedLocPromptOnly{} (range \LocPromptOnlyMin--\LocPromptOnlyMax). Across
\NumLocalizationFaces{} faces, the within-face/model median gain is
\MedPairedLocGain{} (95\% face-clustered bootstrap interval
\PairedLocGainLow--\PairedLocGainHigh). The median reduction in measured
off-target change is \MedPairedOffReduction{} (interval
\PairedOffReductionLow--\PairedOffReductionHigh).

The component values give the same descriptive result: median
$\Delta E_{\mathrm{off}}$ is \MedOffPromptOnly{} for prompt-only outputs and
\MedOffComposite{} after compositing, while median $\Delta E_{\mathrm{tgt}}$ is
\MedTgtPromptOnly{} and \MedTgtComposite{}, respectively. Prompt instructions
alone did not preserve the measured keep zone. Post-hoc compositing did so
because preservation is enforced by the blend; this result does not assess
anatomical correctness inside the mask or changes outside the limited keep zone.

\subsection{Identity and edit strength}
\label{sec:identity}

\begin{figure*}[t]
  \centering
  \includegraphics[width=0.92\textwidth]{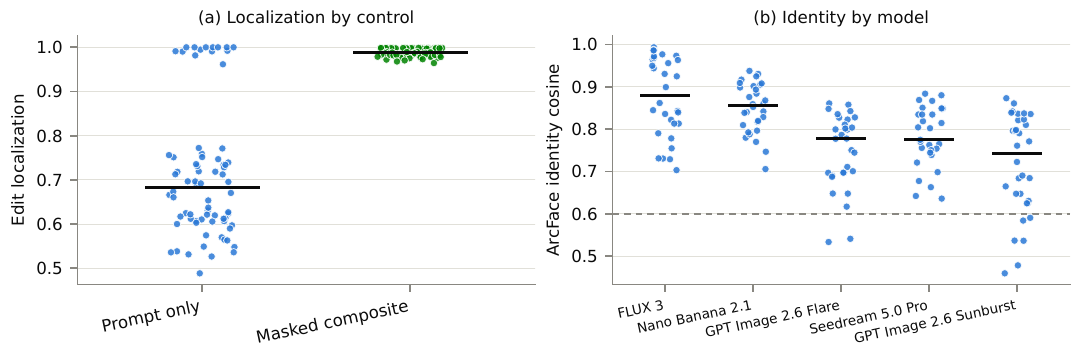}
  \caption{(a) Localization by control, models pooled. (b) Identity by model,
  controls pooled; the dashed line is the heuristic 0.6 reference. Each point is
  one scoreable output (horizontal bar: median); points are repeated outputs,
  not independent patient samples.}
  \label{fig:strips}
\end{figure*}

Prompt-only identity and target-region change trade off across editors
(Fig.~\ref{fig:strips}b). GPT Image~2.6 Sunburst edits most strongly (median
prompt-only $\Delta E_{\mathrm{tgt}}$ \MedTgtFaceliftPromptOnlySunburst{} on the
facelift-style request and \MedTgtRhinoPromptOnlySunburst{} on rhinoplasty) and
has the lowest prompt-only identity (\MedIdentityPromptOnlySunburst); its Flare
tier is close behind (\MedIdentityPromptOnlyFlare). Using 0.6 as a heuristic
reference, \NumBelowFloor{} scoreable outputs fall below it: \NumBelowFloorSunburst{}
from Sunburst and \NumBelowFloorFlare{} from Flare, all on rhinoplasty, with a
minimum of \IdentityMinAll{}. Seedream~5.0 Pro sits in the middle
(\MedIdentityPromptOnlySeedream). Nano Banana~2.1 retains identity best among
editors that perform both edits (\MedIdentityPromptOnlyNanoBanana), with the
smallest facelift-style change of that group
(\MedTgtFaceliftPromptOnlyNanoBanana).

FLUX~3 has the highest prompt-only identity (\MedIdentityPromptOnlyFlux), but on
the facelift-style request it mostly returned the input unchanged. Its median
target change there is \MedTgtFaceliftPromptOnlyFlux{}, compared with
\MedTgtFaceliftPromptOnlySeedream--\MedTgtFaceliftPromptOnlySunburst{} for
Seedream and the two GPT Image tiers, so its localization ratio near 1 reflects
almost no change anywhere. On rhinoplasty it does edit
(\MedTgtRhinoPromptOnlyFlux) and leaves the keep zone nearly untouched. High
localization and high identity together therefore do not show that an edit was
made; the component values are needed to tell the two cases apart.

\begin{figure*}[t]
  \centering
  \includegraphics[width=0.99\textwidth]{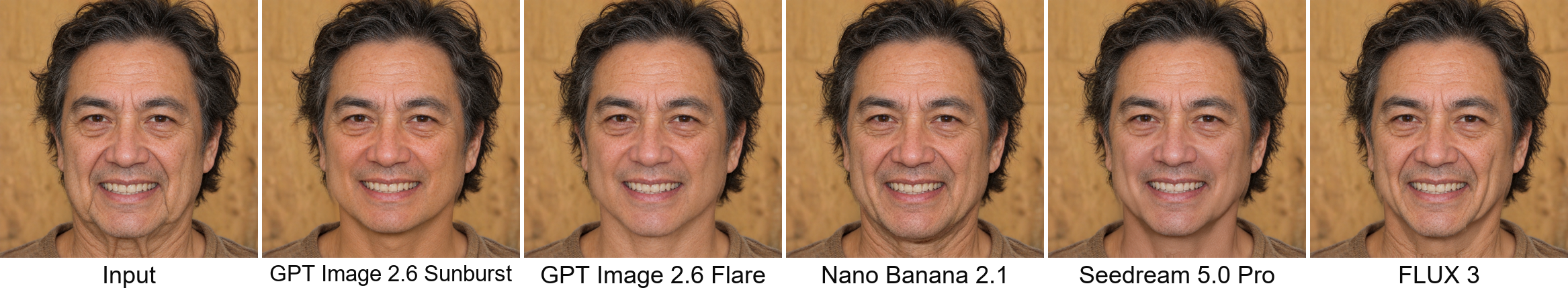}
  \caption{One facelift-style jaw--neck edit of a synthetic face, composited from
  every editor. The visible edit-strength spectrum parallels the automated
  scores, including the near-copy from FLUX~3; the figure is qualitative and was
  not surgeon rated.}
  \label{fig:modelstrip}
\end{figure*}

Fig.~\ref{fig:modelstrip} shows the same face after compositing outputs from each
editor. Prompt-only identity is the model-level diagnostic (median
\MedIdentityPromptOnly{}); composite identity is a system-level property (median
\MedIdentityComposite{}) partly determined by copied pixels. Neither ordering is
interpreted as clinical quality.

\subsection{Chained feasibility probe}

On one face, both chaining orders produced scoreable outputs: identity
\ChainedIdentityLow--\ChainedIdentityHigh{} and localization
\ChainedLocLow--\ChainedLocHigh{} ($n=\NumChained$). No visible interference was
noted in these two inspected outputs. They establish technical feasibility only
and do not support a general order-effect conclusion.

\subsection{Qualitative observations and failures}
\label{sec:failures}

\begin{figure*}[t]
  \centering
  \includegraphics[width=0.88\textwidth]{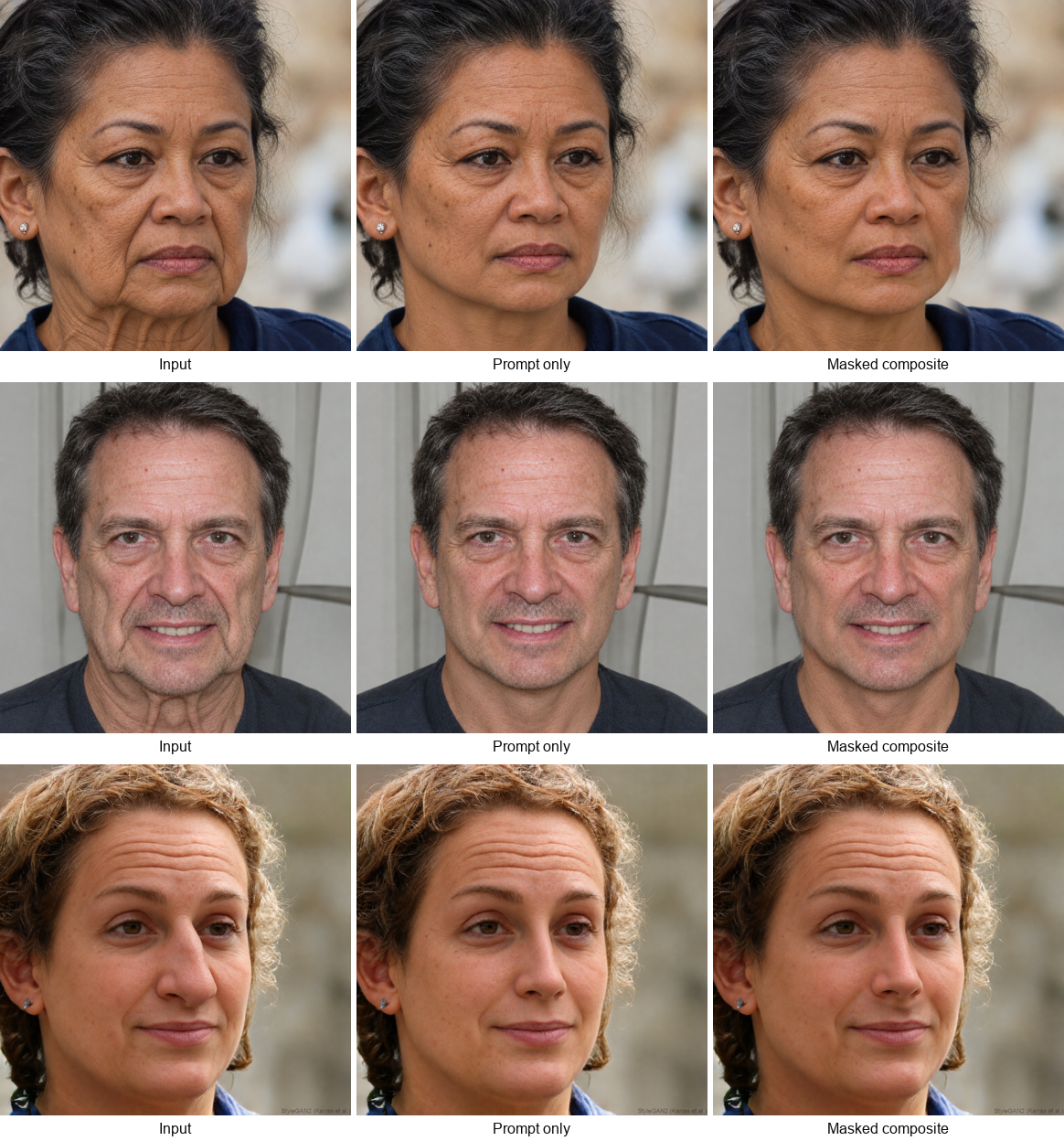}
  \caption{Selected outputs from GPT Image~2.6 Sunburst (top to bottom: two
  facelift-style faces and one rhinoplasty face). Prompt-only outputs alter
  regions outside the request, such as forehead and eye-area wrinkles.
  Composited outputs are unchanged outside the blend mask by design; in the top
  row the original jaw--neck outline shows faintly through the blend at the
  right side of the neck.}
  \label{fig:grid}
\end{figure*}

Figs.~\ref{fig:teaser} and \ref{fig:grid} show selected outputs. The
facelift-style examples contain visible jawline, jowl, and neck-contour changes;
prompt-only outputs also smooth forehead and periorbital wrinkles, which the
composite restores from the input. These appearances were not clinically rated.
Three failures were observed. First, when the editor narrows the jaw--neck contour, the
original outline can show through the feathered blend as a faint ghost contour
(Fig.~\ref{fig:grid}, top row, right side of the neck). Second, FLUX~3
returned near-copies for the facelift-style request, which compositing cannot
repair. Third, the identity detector found no face in one rhinoplasty input,
removing all of its cells, and in one prompt-only output; these account for the
\NumExcludedEdits{} attempted but unscoreable outputs.

\section{Limitations}
\label{sec:limitations}

This is a small pilot: \NumPrimaryFaces{} faces support the main comparisons,
each model--control cell has fewer than ten faces, and the chained probe uses one
additional input. Multiple outputs from one face are correlated, and the two
procedure groups share the same eight synthetic identities; the clustered
interval addresses that dependence descriptively but cannot provide broad
population inference. Each cell contains one generated output, and only one of
the five editors accepts the supplied seed, so stochastic variation is unknown.

The faces are synthetic. They carry no real anatomy, and the procedure-specific
inputs are themselves instruction edits made with GPT Image~2.6 Sunburst, one of
the editors under test. That editor still scored lowest on identity, but the
shared provenance is a possible confound, and real preoperative photographs may
behave differently. The study has no postoperative photographs and therefore
no outcome reference.

The localization ratio measures CIELAB pixel change, not anatomical direction,
clinical plausibility, realism, or expected postoperative appearance. Its keep
zone covers only part of the face, and it abstains on profile views, precisely
where rhinoplasty visualization is important; compositing guarantees off-mask
preservation, so that result verifies system behavior rather than model quality.
ArcFace's 0.6 line is heuristic. Provider-side masked inpainting was not
evaluated: in a pilot attempt, the tested editor's mask input filled the
facelift-style mask region with black, and we excluded it rather than report
artifacts as edits.

No surgeon ratings were collected. Clinical plausibility therefore rests neither
on expert judgment nor on a clinically validated automated score. The face set
spans several apparent ages, sexes, and skin tones but was not demographically
characterized, so the results cannot be generalized across groups; commercial
face pipelines have documented demographic
disparities~\cite{buolamwini2018gendershades}.

Finally, provider endpoints and aliases can change without notice. Run dates and
endpoint identifiers aid auditing but do not pin every underlying model revision.
The public repository contains the canonical score table and paper analysis, not
the complete generation/scoring implementation or input images, so the study is
not presently reproducible end to end.

\section{Ethics and Privacy}
\label{sec:ethics}

\textbf{Synthetic faces.} Every face in the study is synthetic: the identities
were sampled from a generative face model and the procedure-specific inputs were
derived from them by instruction edits. No photograph of a real person, and no
clinical record, was used or sent to any provider. Images were transmitted to the
gateway (fal.ai) inline and outputs returned inline (\texttt{sync\_mode}), so no
stored media objects were created at the gateway.

\textbf{Accounting.} The design attempted every cell in the specified matrix.
Table~\ref{tab:main} contains every scoreable output, and detector failures are
reported explicitly rather than silently omitted. Qualitative figures are
selected for legibility and are not evidence of prevalence; the canonical score
table supports quantitative auditing.

\textbf{Intended use and misuse.} A preview is an expectation-setting aid, not a
promised outcome. Over-idealized previews can create expectations that surgery
cannot meet~\cite{agarwal2007morph}. The present system has not been clinically
validated and should not guide treatment or be shown as a predicted outcome. The
same editing capability applied to a non-consenting person's photograph is
harmful. Any deployment should verify subject authorization, disclose synthetic
content and uncertainty, retain the original image, and log the endpoint and
settings. Model-card documentation practices~\cite{mitchell2019modelcards} apply
to systems assembled from commercial APIs as well as to their component models.

\textbf{Fairness.} The demographic makeup of the test set was not characterized,
and face analysis systems can exhibit demographic performance
gaps~\cite{buolamwini2018gendershades}. A stratified evaluation on consented
real photographs, with failure analysis, is a prerequisite for any clinical-use
claim.

\section{Conclusion}
\label{sec:conclusion}

In this pilot, text instructions alone did not prevent current image-editing
APIs from changing a predefined facial keep zone. A landmark-derived mask,
alignment, and client-side feathered composite reduced that measured off-target
change while retaining most of the target-region pixel change, uniformly across
the five editors tested and at a median provider cost of \$\MedCost{} per image.
Reporting target and off-target change separately also exposed an editor that
returned near-copies of the facelift-style input, which a high identity score
alone would have hidden.

The study does not validate cosmetic-surgery outcome prediction, and client-side
compositing should be read as a control for off-target pixel changes, not as
evidence that a preview is clinically accurate or safe for patient counseling.
The main analysis contains \NumPrimaryFaces{} synthetic faces, one generated
output per cell, no surgeon ratings, no postoperative reference, and limited
profile support. The next study should use consented real preoperative and
postoperative photographs, add repeated generations, pose-robust whole-image
localization, and anatomy-aware accuracy metrics. It will also need multiple
blinded surgeons with inter-rater agreement, a larger and demographically
characterized cohort, and immutable or self-hosted model baselines. What the
pilot does establish is narrower: when an editor exposes no internals, region
confinement can still be enforced entirely from the client.

\bibliographystyle{IEEEtran}
\bibliography{refs}

\end{document}